\documentclass[letterpaper, 10 pt, conference]{ieeeconf}  

\IEEEoverridecommandlockouts                              

\usepackage{amsmath} 
\usepackage{amssymb}  
\usepackage{graphicx}
\usepackage{caption} 
\usepackage{algorithm}
\usepackage{algorithmic}
\usepackage{flushend}
\usepackage{makecell}
\usepackage{color}
\usepackage{overpic}
\usepackage{soul}
\usepackage{csquotes}
\usepackage{multirow}
\usepackage{multicol}
\usepackage{wrapfig}
\usepackage{booktabs}
\usepackage{threeparttable}
\usepackage{hyperref}
\usepackage{pifont}

\newtheorem{theorem}{Theorem}
\newtheorem{definition}{Definition}

\usepackage{graphicx}
\usepackage{subfig}
\usepackage{lipsum}
\usepackage{afterpage}

\makeatletter
\newcommand{\setcaptype}[1]{\def\@captype{#1}}
\makeatother

\newsavebox{\tempbox}

\title{Imagined Potential Games: A Framework for Simulating, Learning and Evaluating Interactive Behaviors}

\author{Lingfeng Sun$^{1}$ \quad Yixiao Wang$^{1}$ \quad Pin-Yun Hung$^{1}$ \quad Changhao Wang$^{1}$ \quad Xiang Zhang $^{1}$ 
\\ Zhuo Xu$^{2}$ \quad Masayoshi Tomizuka$^{1}$ 
\thanks{$^{1}$Department of Mechanical Engineering, 
        University of California, Berkeley, Berkeley, CA 94720, USA.
        {\tt\small \{lingfengsun, tomizuka\}@berkeley.edu}}%
\thanks{$^{2}$Google Deepmind, Mountain View, CA 94043, USA.
        {\tt\small zhuoxu@google.com}}%
}

\begin{document}
\maketitle
\savebox{\tempbox}{\begin{minipage}{\textwidth}
\setcaptype{figure}%
\centering
    \includegraphics[width=\textwidth]{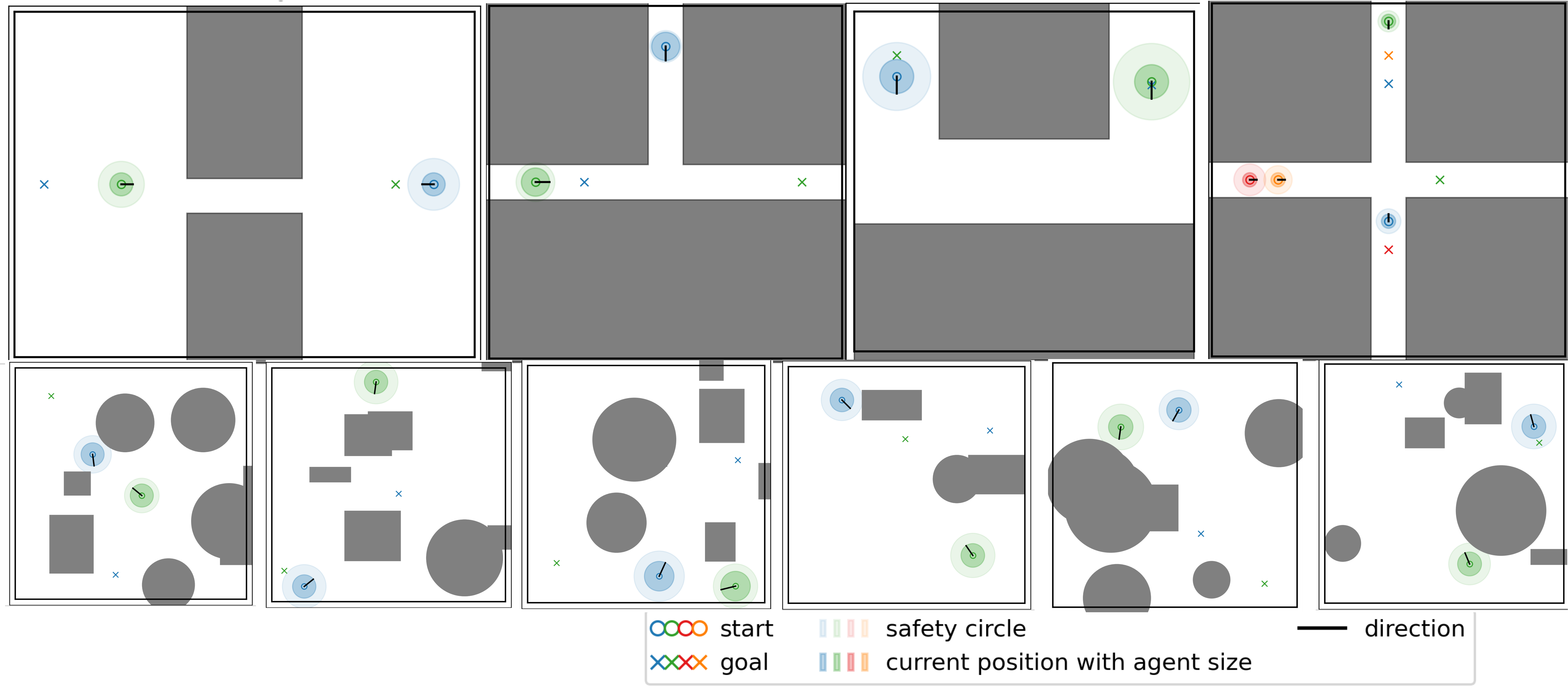}
    \caption{Selected interactive scenarios: Hallway, U-turn, T-intersection, Intersection, and randomly generated interactive scenarios. Vanilla collision-avoidance planners might not work in these scenarios, and agents must collaborate to work out cooperative solutions. We simulate interactions in these scenarios and use them to learn and evaluate interactions.}
    \label{fig:teaser}
    \vspace{-0.2in}
\end{minipage}}
\begin{figure}[t]
\rlap{\usebox\tempbox}
\end{figure}
\afterpage{\begin{figure}[t]
\rule{0pt}{\dimexpr \ht\tempbox+\dp\tempbox}
\end{figure}}

\begin{abstract}
Interacting with human agents in complex scenarios presents a significant challenge for robotic navigation, particularly in environments that necessitate both collision avoidance and collaborative interaction, such as indoor spaces. Unlike static or predictably moving obstacles, human behavior is inherently complex and unpredictable, stemming from dynamic interactions with other agents. Existing simulation tools frequently fail to adequately model such reactive and collaborative behaviors, impeding the development and evaluation of robust social navigation strategies.
This paper introduces a novel framework utilizing distributed potential games to simulate human-like interactions in highly interactive scenarios. Within this framework, each agent imagines a virtual cooperative game with others based on its estimation. We demonstrate this formulation can facilitate the generation of diverse and realistic interaction patterns in a configurable manner across various scenarios.
Additionally, we have developed a gym-like environment leveraging our interactive agent model to facilitate the learning and evaluation of interactive navigation algorithms. 
Results and code are available on the project website: \href{https://sites.google.com/view/simulate-learn-interact}{https://sites.google.com/view/simulate-learn-interact}.
\end{abstract}


\section{Introduction}
Developing effective navigation policies is essential for enhancing human-robot interaction. A critical component of this development is the creation of diverse, interactive agents capable of mimicking human-like behaviors. However, simulating realistic human interaction within autonomous systems presents significant challenges. Current simulation models primarily focus on simple collision avoidance, which is often inadequate for complex interactive scenarios. For instance, in a narrow hallway where only one person can pass at a time (Figure~\ref{fig:teaser} upper left), mere collision avoidance is insufficient. Such scenarios require collaborative behaviors, such as one individual stepping back to allow another to pass, thereby preventing a deadlock. Moreover, the system must also identify optimal moments to yield when others are advancing, mirroring the implicit and non-verbal coordination of movements observed in real human interactions.
In these settings, the agent must interact with others in a closed-loop manner. If the simulated agent is overly conservative, the resulting navigation policies might become aggressive, potentially raising safety concerns. Conversely, overly conservative policies could cause the agent to get stuck in crowded environments. 

To address these issues, the primary challenge lies in generating or simulating diverse yet realistic behaviors in a closed-loop manner that can effectively interact with other agents. This involves understanding, predicting, and dynamically adapting to the complex interplay of multiple agents within shared spaces. One method is to collect extensive interaction datasets to train interactive behavior policies. However, collecting large amounts of human-interacting data across multiple scenarios is expensive. And even in collected human motion datasets, interactions are typically rare and tailed events. In the autonomous driving domain, using the collected motion dataset, the state-of-the-art autonomous driving simulator Waymax~\cite{gulino2023waymax} employs data-driven prediction models and rule-based models that follow recorded paths as closed-loop planners for reactive agents. However, these methods are not designed for closed-loop interactions and unsuitable for highly interactive cases as they are prone to failures in situations not well represented in the data, such as sudden, out-of-distribution changes during an interaction.
\begin{figure*}[h!]
    \centering
    \includegraphics[width=\textwidth]{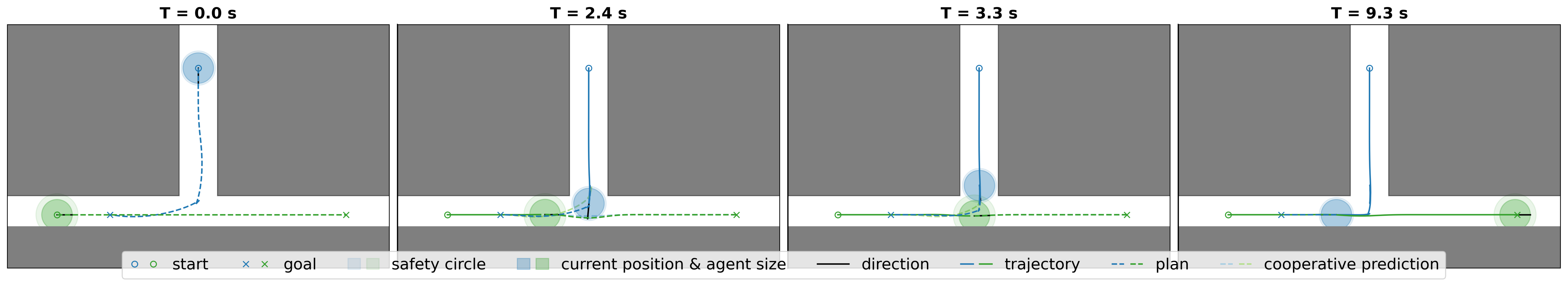}
    \vspace{-0.2in}
    \caption{A human-like interaction at the T-intersection: one agent stepped back to yield.}
    \label{fig:t-intersect}
\vspace{-0.18in}
\end{figure*}
This paper aims to explore the simulation of human-like interactions in a distributed setting and the use of such simulations to enhance the learning of collaborative interaction strategies. 
The distributed setting represents the practical case in interactions, where each agent independently formulates plans based on its observations without access to the plans or cost function parameters of other agents. Figure \ref{fig:teaser} illustrates some challenging indoor navigation scenarios where collaborative planning is necessary under the presented starting/goal configurations, and figure \ref{fig:t-intersect} shows an illustrative interaction trajectory generated from our proposed method at a T-intersection. One agent moved back to yield after coming out of the intersection and saw the other agent coming through. It is worth noticing that the behavior emerges from closed-loop simulation without predefined intention or motion.

In this paper, we first introduce a framework based on distributed imagined potential games (IPG) that can simulate interactions in scenarios where traditional collision-avoidance algorithms fail. Potential games provide a method for deriving game-theoretical equilibrium plans for multiple agents. We further extend this concept to a distributed IPG, where each agent independently imagines a potential game with others based on its estimations. We find this distributed IPG formulation is crucial for achieving diverse and realistic interactive behaviors akin to human interactions. We employ trajax~\cite{trajax} to implement the fast online iLQR optimization~\cite{pilqr} and verify such a framework can generate diverse and realistic interactions effectively in various scenarios shown in Figure \ref{fig:teaser}. Furthermore, We integrate these reactive distributed agents into a gym-like environment, enabling the simulation of interactions within both provided and randomly generated settings. This environment serves as a platform for training reinforcement learning agents and evaluating navigation algorithms designed for interaction. Finally, we discuss the ongoing challenges in developing metrics to evaluate interaction generation in new scenarios and the capability of training navigation policies to collaborate with various reactive agents using reinforcement learning.


\section{Preliminaries and Background}
\subsection{Distributed Multi-agent Planning}
Assume we have $N$ agents in the scenario. For each agent $i, 1\leq i\leq N$, let the vector $x_i(t)\in \mathbb{R}^{n_i}$  denote the state of agent $i$, let $u_i(t)\in \mathbb{R}^{m_i}$ denote the control input of agent $i$ at time $t$. Each agent follows its system dynamics $x_i(k+1)=f_i(x_i(k), u_i(k))$.
Unless otherwise specified, throughout this paper, for variable $x$, we use a subscript \(x_i\) to denote agent $i$ and a superscript \(x^k\) or \(x(k)\) to indicate the time horizon $k$. If $i$ and $k$ are not specified, it means $x$ for all agents or all time steps. $x_{-i}$ denotes $x$ of all agents excluding $i$. 

Obstacles in the scenario are represented by $\{O_j\}_{j=1}^M$. Each agent has its initial state $x_i^0$ and a target goal state in the scenario $g_i$. All the agents in the scenario navigate to their target goal while avoiding collision with the environmental obstacles and other agents. Interactions happen when their planned trajectories $\{x_i(0), x_i(1),...,x_i(T)\}_{i=1}^N$ have conflicts and need to interact to reach non-conflict new plans. Control inputs of multiple agents $U=[u_1^{0:T},u_2^{0:T},...,u_N^{0:T}]$ are the plans of all the agents.
Under the \textit{distributed setting with no communication}, we assume each agent $i$ is solving an optimal control optimization without knowing others' plans.
\begin{equation}
\begin{aligned}
&\min_{u_i(0:T), x_i(0:T)} \quad  J_i(x(0), u_i,\Tilde{u}_{-i})\\
&\textrm{s.t.} \quad 
x_i(k+1) = f_i(x_i(k), u_i(k)), \quad h(x_i,\Tilde{x}_{-i}, O) \leq 0\\
\end{aligned}
\label{eq:distributed_planning}
\end{equation}

$J_i=S_i(x(T), T) + \sum_{k=0}^{T-1}L_i(x(k), \Tilde{u}_{-i})$ is the cost function for agent $i$. The running cost $L_i$ include distance, time, and energy costs, and the terminal cost $S_i$ can include goal conditions. The collision-free requirements are in constraints $h\leq 0$. The hard constraints in $h$ can be added as weighted cost functions depending on the solver used. 

\textbf{Collaborative-prediction required interactions} 
$\Tilde{u}_{-i}$ and $\Tilde{x}_{-i}$ are the estimation of other agents' plans and states to prevent collisions since we assume no communication. In single-agent navigation with obstacle avoidance or open-area social navigation for multiple agents, constant velocity predictions with model predictive control can provide good local collision avoidance planning~\cite{van2011reciprocal}. However, in highly interactive scenarios shown in Figure \ref{fig:teaser}, with some initial and target positions, there are no feasible collision-free plans for each interaction agent. Online collision avoidance algorithms are insufficient; collaborative predictions are required.

\textbf{Comparison with the \textit{centralized} or \textit{distributed with sharing} setting:} 
In the \textit{centralized} setting, $U$ is solved together in a single large problem given all agents' initial and goal states simultaneously. The weighted costs of all agents are optimized in one problem.
The \textit{distributed setting with sharing} is quite similar to the centralized setting; plans are solved separately but shared with other agents, enabling accurate predictions for cooperation in a distributed setting \cite{luis2020online}. Many game-theoretical interaction models operate in a similar setting. The cost functions of agents are shared to consider others' behaviors and find equilibrium plans for all. 

\subsection{Potential Game}

We follow the dynamic games ($N$ agents with horizon $T$) settings introduced in ~\cite{talha2021potential,alejandra2018potential,guo2023markov,guo2024alpha}.
A differential game is described by the compact notation of $\Gamma_{x_0}^T=(N, \{U_i\}_{i=1}^N, \{J_i\}_{i=1}^N, \{f_i\}_{i=1}^N)$, where $x_0$ is the initial states of all agents, and each agent seeks to optimize its cost $J_i$ under the dynamic $f_i$. The cost function $J_i(x(0), U)=S_i(x(T)) + \sum_{k=0}^{T-1}L_i(x(k), U(k))$ consists of running cost $L_i$ and terminal cost $S_i$. We look for the Nash equilibrium solution of the dynamic game. At a Nash equilibrium, no agent has the incentive to change its current control input $u_i^*$ as such a change would not yield any benefits, given that all other agents' controls $u_{-i}^*$ remain fixed. While this equilibrium solution can best represent the cooperative multi-agent behavior in the interaction, finding the Nash equilibrium solution is challenging since there are $N$ coupled optimal control problems to solve simultaneously. \cite{talha2021potential} proved that a centralized optimal control using iLQR can solve the potential differential game form, allowing us to solve for the equilibrium online. Due to page limits, we leave formulation details in Appendix A on the website\cite{website}.

\begin{figure*}[t]
    \centering
    \includegraphics[width=\textwidth]{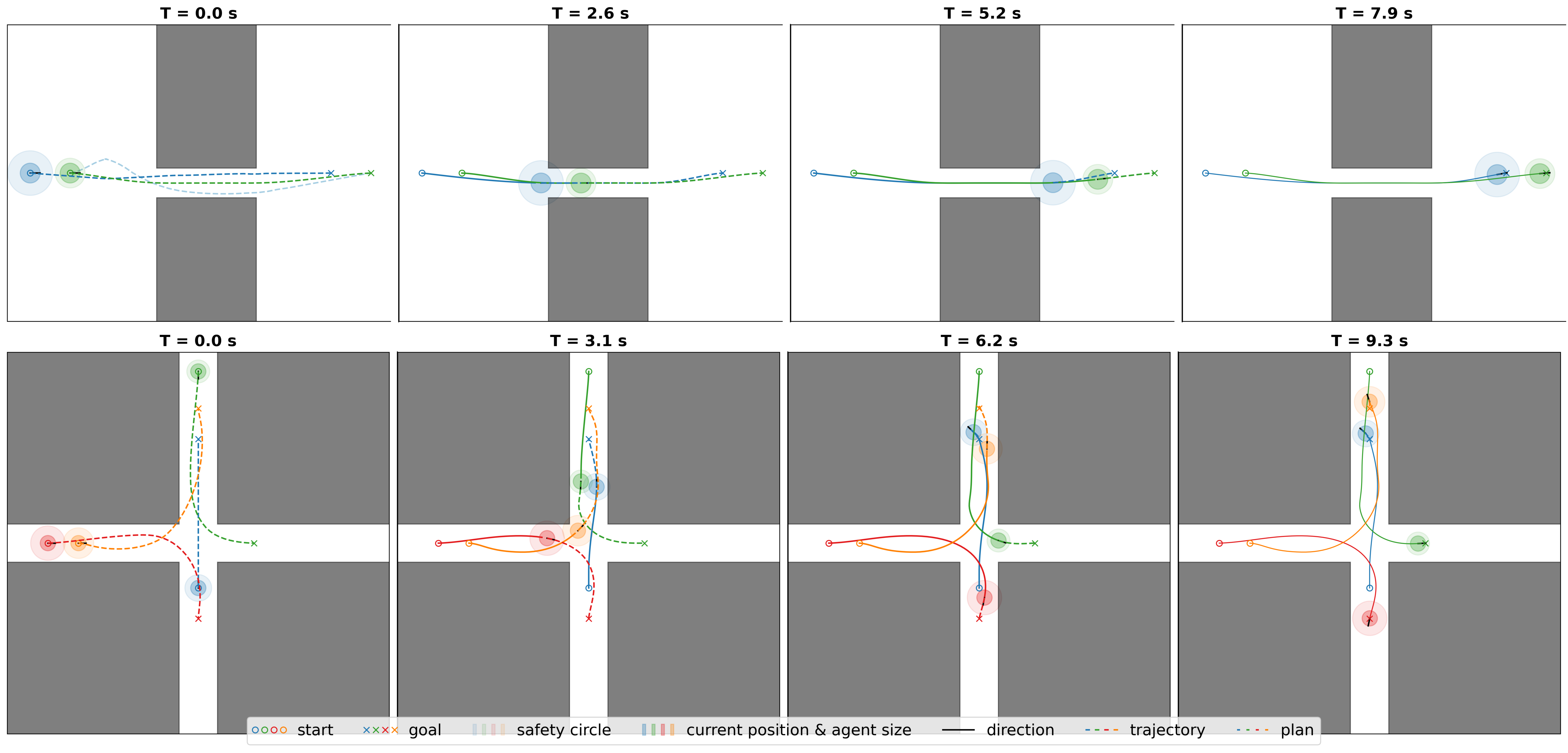}
    \vspace{-0.2in}
    \caption{Selected interaction trajectories from the generated closed-loop interactions in Hallway and Intersection scenarios using randomized configurations. Cooperative predictions are hidden for clearness in interactions of more than two agents.}
    \label{fig:ipg_result}
\vspace{-0.18in}
\end{figure*}

\section{Imagined Potential Game for Collaborative Interaction Generation}
Equation~\ref{eq:distributed_planning} outlines the distributed framework for agent interactions without communication. Within this framework, each ego agent needs to estimate the future behaviors of other agents $(\Tilde{x}_{-i}, \Tilde{u}_{-i})$ to facilitate collision avoidance.
This distributed formulation serves as a general model applicable to both indoor navigation and autonomous driving, mirroring how humans plan their actions based on their estimations of others' behaviors. However, a significant challenge in applying this formulation to determine the ego agent's behavior is accurately assessing the intentions of other agents. To address this, we employ an Imagined Potential Game (IPG) approach.

The Imagined Potential Game~\cite{sun2024distributed} formulation enables cooperative predictions about the behaviors of other agents and assists in formulating the current plan for the ego-agent. In this model, the ego-agent anticipates the actions of other agents by imagining a potential game among them. To effectively simulate this game, it needs to know the goal position and interaction parameters of other agents. We assume other agents' long-term goal positions (intentions) are already given since this project focuses more on short-horizon local interaction rather than long-term goal prediction. In practice, state-of-the-art goal prediction networks can be used to predict goals. For interaction parameters, agents start assuming other agents using the same parameters they have, e.g., $\Tilde{r}_j=r_i$ for all $j\neq i$. This assumption is simple but strong enough to simulate diverse interactions in most scenarios, especially when all participants are cooperative agents. For cases where inaccurate estimation causes failures, we introduce methods to break deadlocks in section \ref{sec:ipg:implementation}.

During simulating, $N$ separate potential games exist simultaneously using this formulation in an $N$ agent interaction. The full IPG problem for each agent is shown in Equation \ref{eqn:ipg}, where $p(x, u)$ and $\bar{s}$ are the common terms in the running cost $L_i$ and terminal cost $S_i$ of all agents. $x_U, x_L, u_U, u_L$ are the state and input boundaries, and $r_{obs}$ is the radius of the circle obstacle. Interaction parameters include safety radius $d_i$, and different weights $Q_i, R_i, D_i, B_i$ in cost functions affecting the behaviors. See Appendix A for a detailed definition of interaction parameters.

\begin{minipage}{0.45\textwidth}
        \begin{algorithm}[H]
            \centering
            \caption{Closed-loop Distributed IPG}
            \label{alg:algo1}
                    \begin{algorithmic}
                        \STATE {\bfseries Initialize:} U = 0, $\{\tau_i\}_{i=0}^N$ = empty \\
                        \WHILE{termination condition not satisfied}
                         \STATE {\bfseries for} i in N {\bfseries do in parallel} \\
                         \STATE $\quad U \leftarrow iLQR(x_0, g, Q_i, R_i, D_i, B_i, d_{i})$ \\
                         \STATE $\quad u_i^0 \leftarrow U^0_i, x_{next, i} \leftarrow f(x_0, u_i^0)$ \\

                         \STATE {\bfseries for} i in N {\bfseries do} \\
                         \STATE $\quad x_{0, i} \leftarrow x_{next, i},\tau_i \leftarrow [\tau_i, x_{0, i}]$  \\
                         \STATE $x_{0} \leftarrow x_{0, i=1,...,N}$ (Update initialization)\\
                         
                        \ENDWHILE
                    \end{algorithmic}
        \end{algorithm}
\end{minipage}

\begin{minipage}{0.45\textwidth}
    \begin{equation}
    \label{eqn:ipg}
    \begin{aligned} 
        \min_{U}  \ &\sum_{k=1}^{T-1} p(x(k), u(k)) + \bar{s}(x(T))\\ 
        s.t. \quad &x(k+1) = f(x(k), u(k)), x(0) = x_0 \\ 
        &x_L \leq x(k) \leq x_U, u_L \leq u(k) \leq u_U \\ 
        &r_{obs} - dis(x(k), O_m) \leq 0, m=1...M \\ 
        &r_{i} - dis(x_i(k), x_j(k)) \leq 0, i, j=1...N\\
    \end{aligned}
    \end{equation}
\end{minipage}

Algorithm~\ref{alg:algo1} outlines the process of closed-loop simulation of the interaction: each agent solves the IPG using iLQR. Following the receding horizon control manner, only the first step it solved for its own game is used for itself, denoted as $U_i^0$ is used. All agents can solve their IPG in parallel. The stored trajectories $\{\tau_i\}_{i=0}^N$ are the simulated closed-loop interactions. Below, we report challenges met to get rapid and stable simulations in certain cases. 

\subsection{Improvements on Interaction generation for vanilla IPG}
\label{sec:ipg:implementation}
\textbf{Increasing number of agents:}
The optimization problem described in Equation~\ref{eqn:ipg} becomes harder to solve when the number of agents and obstacles increases, e.g., more sensitive to initialization and takes longer. The agent number in a fixed environment can vary a lot. To keep the interaction generation efficient, we assume that each agent only interacts with the closest $N=3$ agents around.

\textbf{Obstacle shape:}
We extend the circle obstacle shape in the previous IPG framework~\cite{sun2024distributed} to circle and rectangle shapes for more flexible environment designs. Square obstacles are included using the soft constraints by adding the collision penalty into the cost using a distance function penalizing the distance to the two closest sides. See details in the appendix.

\textbf{Occlusion:} 
\label{sec:ipg:occlusion}
While the naive version of the IPG framework assumes the full observability of other agents, occlusion has huge effects on interactions. Therefore, other agents are excluded from the potential game when they are behind the ego agent, obstacles exist on the line connecting the agents, or the agents are out of a pre-defined observation range.

\textbf{Solving deadlocks:} Deadlocks
Deadlock cannot be completely avoided under the distributed framework; it frequently happens when occlusions exist (e.g., two agents fail to interact well when one observes a third agent, but the other cannot) or under some initial/goal position configurations(e.g., identical agents in symmetric positions).
We randomly sample the agent parameters during interaction simulation to avoid identical agents. When a deadlock is detected during the closed-loop generation, we implemented two methods for a randomly selected agent: 
\emph{1)} Increase the ego agent's safety distance to change interaction style. The insight is that safety distance is important in generating diverse behaviors such as yield, cut-in, etc. \cite{sun2024distributed}. 
\emph{2)} Aggressive planning for the selected agents by changing the estimated goal points of other agents to their current position--ego assumes others will always yield to cooperate.

The framework aims to generate interactions in new environments with arbitrary agent numbers and initial conditions. See more details on implementation in Appendix B.

\subsection{Results}
\label{sec:ipg:results}
\textbf{Interaction in various scenarios:}
In Figure~\ref{fig:ipg_result}, we show the interaction generated using randomly sampled agent parameters and initial/goal points. We select the keyframes in the interactions to demonstrate the yielding behaviors that naturally appeared in the simulation. 
We present both the planned trajectories and the predicted collaborative trajectories of other agents to illustrate the different potential games being solved during two-agent interactions.

\textbf{Baselines Comparison:}
We compare our method to ORCA~\cite{orca}, a strong baseline for distributed collision avoidance. We implemented ORCA using a global reference trajectory solved by single-agent trajectory optimization (same optimization as IPG, but neglecting all other agents). ``Success’’ means the agents can reach goals in a limited time. ``Collision’’ means the agents collide with other agents or the environment during simulation. ``Timeout’’ means the agents fail to reach goals within the time limit (deadlock happens or is solved too slowly). We show results in Table \ref{tab:orca_compare} at the narrow hallway(H) and T-intersection(T) using randomized configs. Comparisons are shown on the website~\cite{website}.
\begin{table}[h]
    \vspace{-0.15in}
    \caption{Evaluation in Hallway and T Scenarios}
    \vspace{-0.05in}
    \centering
    \begin{tabular}{l|llll}
\Xhline{1pt}
               & IPG-H & ORCA-H & IPG-T & ORCA-T  \\ 
    \hline
    Success    & 20/20  & 14/20 &  19/20 & 11/20\\
    Collision  & 0/20  & 0/20 &  0/20 & 0/20   \\
    Timeout    & 0/20 &  6/20 &  1/20 & 9/20  \\
\Xhline{1pt}
\end{tabular}
    \vspace{-0.1in}
    \label{tab:orca_compare}
\end{table}

\textbf{Controllable generation:}
Different parameters in the optimizations represent the characteristics of agents in the interaction. The change of parameters influences agent behaviors in interactions. Adjusting the parameters can result in different types of agents and interactions. A detailed study of the effects of different parameters on interaction behavior and solver stability can be found in Appendix B.

\begin{figure}[h]
    \centering
    \vspace{-0.1in}
    \includegraphics[width=0.85\linewidth]{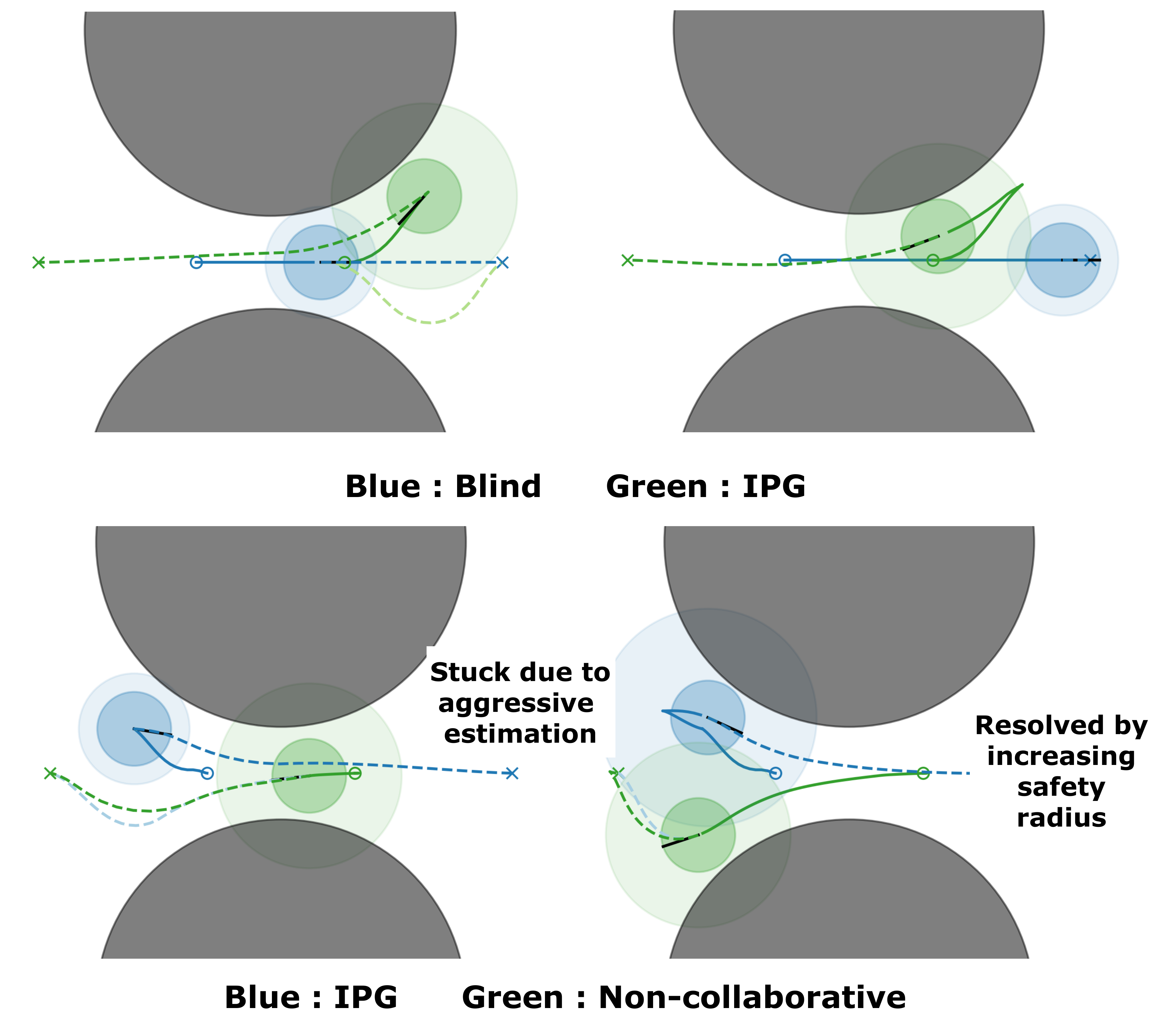}
    \vspace{-0.1in}
    \caption{IPG agents interact with different types of agents.}
    \label{fig:ipg_heterogeneous}
\vspace{-0.15in}
\end{figure}

\textbf{IPG interacting with heterogeneous agents:}
\label{sec:ipg:heterogeneous}
The distributed nature of the framework enables IPG agents to engage with any other agent and the closed-loop planning mechanism ensures safe interactions.
Figure~\ref{fig:ipg_heterogeneous} illustrates the IPG agent's interactions with two different types of agents. In the top scenario, the IPG agent interacts with a blind agent that ignores others. In the bottom, the IPG agent engages with a non-collaborative agent, focusing solely on collision avoidance.
Notably, the IPG agent's flexible behavior allows it to adapt to unexpected situations. For instance, when the IPG agent gets stuck due to the green agent's reluctance to move, increasing the safety distance resolves the deadlock by aligning the green agent's cautious behavior. This enables the IPG agent to navigate the situation effectively.

The experiments above show that the IPG agent can interact with different kinds of agents without knowing their plans in advance. Therefore, we can use them as intelligent planners in collaboration-required environments in simulation and the real world.
Currently, we only include 2D trajectories in indoor scenarios to demonstrate the collaborative planning and interaction generation capabilities. Experimenting on mobile robots with advanced localization, perception, and goal prediction methods integrated is beyond the current scope and left to future work. This distributed setting and behavior allows us to use IPG agents as reactive agents in the environment to train and evaluate RL agents that learn to interact as introduced in Sec~\ref{sec:rl}.

\section{Learning to Interact with Collaborative Agents using IPG}
\label{sec:rl}
Seeing the capabilities of IPG agents interacting with different agents in interactive scenarios, we include them as reactive agents in a simulation environment to build an interactive environment that can train interactive agent policies via reinforcement learning or evaluate an existing interactive navigation policy. The main focus of this section is to show the environment we build and plan to open-source for training/evaluation, demonstrate results using RL baselines, and discuss the challenges for training and evaluation.

\subsection{Environment settings}
\textbf{Scenarios:} Hallway, a T-Intersection, Intersection is implemented for interactive navigation. Each scenario has two configurations: \textit{Standard} and \textit{Wide}. In the \textit{standard} setting, hallways only allow one agent to pass. 

\textbf{Environmental agents:} Agents in the environments have parameters including radius, velocity/acceleration/angular velocity limit, safety distance, and agent type (\textbf{IPG, ORCA, Blind}). Blind agents model extremely aggressive agents that ignore others. We also allow users to customize reactive agents and we are continuously adding more strategies.

\textbf{Training agent:} Unlike in interaction generation, where all agents are IPG, in this environment, one of the agents in the interaction is controlled by a user-provided planner. The control agent has access to environmental information and observed agents' states (see occlusion and observation range in section~\ref{sec:ipg:occlusion}). The control agent outputs the action of acceleration $a$ and angular velocity $\omega$ for the subsequent step. This can be a learned RL actor or any pre-trained policy.

\begin{figure}[t]
\begin{centering}
\includegraphics[width=\linewidth]{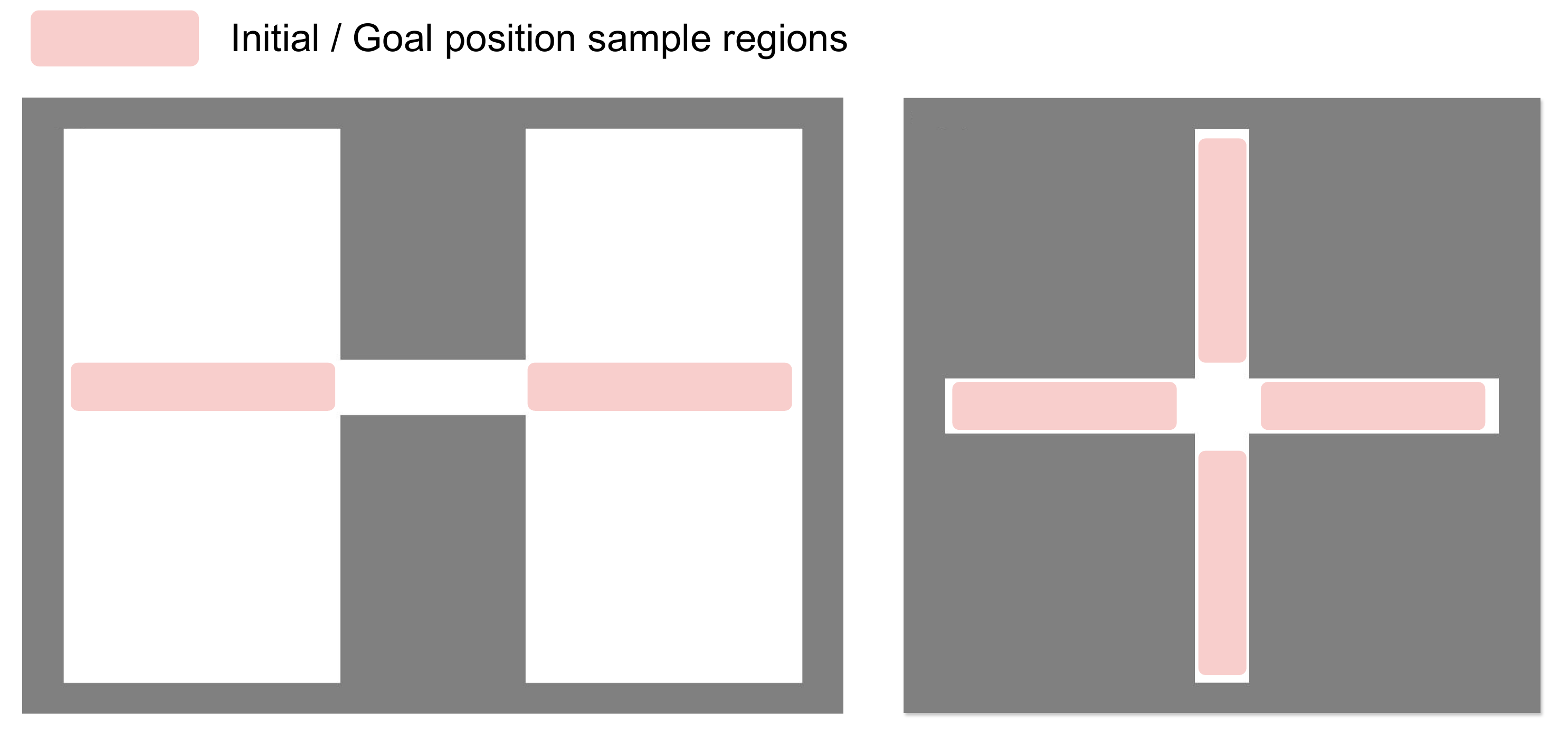}
    
\end{centering}
\vspace{-0.1in}
\caption{Interaction zones in Hallway and Intersection.}
\vspace{-0.1in}
\label{fig:envs}
\end{figure}

\textbf{High Interactive configurations generation:}
Strong interactions only arise when there are conflicts between agents. In social robot navigation tasks, these conflicts typically manifest as overlapping future trajectories. These interactions are tail events in everyday planning, and we've verified that interactions rarely happen between agents using randomly generated starting/goal positions.
As a result, we select zones for potential initial and goal states (initial and goal states are in different zones) to increase the likelihood of trajectory overlaps and randomly generate the configurations. 
\subsection{Reinforcement learning setting and Initial Results}
The ultimate goal of interactive planning is to interact with different types of agents in different scenarios. We allow randomizing scenarios, reactive agent types, and starting/goal positions for training configurations. 
The task is done when the controlled agent reaches the goal point. Collision and exceeding the maximum horizon (stuck) will fail. We choose a standard dense reward setting for navigation, with the goal reaching reward $r_g=||x(t-1)-x_g||^2-||x(t)-x_g||^2$, and the energy reward $r_e=-(|a|+|w|)$. We also have a large terminal reward when reaching the goal.

\begin{table}
\centering
\caption{RL training results in two scenarios.}
\vspace{-0.15in}
\begin{tabular}{cccc}\\\toprule  
\label{table:rl}
Environment &Configuration & IPG & Blind \\\midrule
HallWay &Standard&\ding{52} & \ding{56}\\ 
T-Intersection &Standard&\ding{52} & \ding{52}\\  \midrule
\end{tabular}
\vspace{-0.2in}
\end{table}

\textbf{Baseline initial results:}
Using the environment setting above, we can learn some interacting behaviors (similar to how IPG agents behave), including yielding, using the baseline PPO algorithms for a single scenario with a fixed reactive agent type. It's the simplest setting but non-trivial, especially with the one-pass ``standard'' setting. In Table~\ref{table:rl}, we show the settings where we successfully trained reactive RL agents. Interacting with the IPG agent can be quite easy if the IPG agent is conservative and always yields to the controlled agent.
The difficulty of interacting with a blind agent is that the goal of reaching a reward might discourage the yielding behavior, and the agent takes longer to explore it. Qualitative results, including animation of success and failure cases, can be found on the website\cite{website}.

\textbf{Difficulties in reinforcement learning for navigation:}
While RL policies are closed-loop policies that run fast online, training RL policy in this environment is non-trivial. RL agents are required to react while navigating to the goal. In the training, we find it hard to learn these simultaneously with the current reward terms: it takes long for agents to learn to reach the goal after learning to yield. Such behavior requires either different exploration/training strategies or reward design (see Appendix C for examples of RL agents).
Designing more reliable RL algorithms to learn better interaction behaviors is a challenging task, we hope to train agents react to arbitrary types of agents which could potentially represent different types of human in real-world.

\subsection{Evaluating interaction generation and interactive agents}
\begin{figure}[h]
    \centering
    \includegraphics[width=\linewidth]{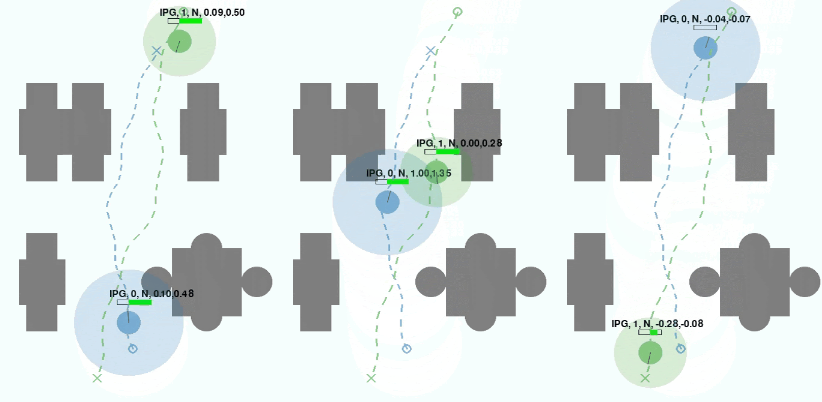}
    \vspace{-0.2in}
    \caption{Reconstructed real interactions (dashed) using IPG.}
    \label{fig:reconstruct}
\vspace{-0.2in}
\end{figure}

Evaluating the quality of simulated interactions is challenging. The most accurate, though inefficient, method is to gather human feedback through questionnaires. Quantitative metrics, while necessary, often fall short of capturing realism. For example, metrics like the extra time taken due to interaction, as discussed in \cite{sun2024distributed}, provide useful data but are insufficient for realism. 
Another helpful realism metric is testing whether the model can reconstruct real interactions. We show how IPG ``reconstructs'' a real motion-captured interaction in Figure~\ref{fig:reconstruct}. Under the same initial position and goal settings, we are able to replay a “similar” interaction behavior by IPG agents with selected parameters. We can also generate different interactions in the same case using different parameters(max velocities, safety radius).
Our implemented environment, with random configurations, serves as a platform for evaluating interactive navigation planners. 
\section{Related Works}
\textbf{Multi-agent planning:}
Multi-robot trajectory planning algorithms can be categorized based on where the computation is done. Two main strategies to solve the problem are \textit{centralized} and \textit{distributed}.
Most previous works use multi-agent path finding (MAPF) solvers with trajectory optimization algorithms~\cite{2016tang, 2020park, 2020michael, wang2022bpomp, wang2021trajectory} to find feasible trajectories for all agents. A series of works ~\cite{2019fisac,2020spica, 2019wang, 2022miller, 2022algames, 2023laine,distributed_pg, 2023mehr} models interactions in multi-agent planning via game theoretic frameworks. 
In the distributed setting, each agent runs a separate algorithm to compute its own trajectory. Depending on whether communication exists between the agents. 
Reactive algorithms~\cite{orca, 2017wang, 2023jian} like Optimal Reciprocal Collision Avoidance (ORCA) can effectively avoid collisions but fail to avoid deadlocks in environments with dense obstacles~\cite{rlss}. Learning-based reactive strategies~\cite{primal, glas, gnn, batra2021} are computationally more efficient but suffer from distribution shifts and also experience deadlocks.
dMPC~\cite{dmpc} and MADER~\cite{mader} consider longer horizons and generate sequences instead of single actions but require communication for collision avoidance.

\textbf{Interaction generation and interactive simulators:}
Simulating the behaviors of actors is an important task with a wide range of applications in transportation and robotics research, where simulators play essential roles in helping train and evaluate intelligent agents.
Many autonomous-driving research works focus on simulating distributed agents' behaviors in the simulator to generate realistic interactions and reactive agents~\cite{Yeh_2019_CVPR, Suo_2021_CVPR, 2021diverse, 2023editing, 2022trajgen} or analyzing and predicting interactive behaviors of heterogeneous agents\cite{qcnet, 2022pseudo, 2019itsc}.
Most autonomous driving simulators~\cite{gulino2023waymax, Dosovitskiy17carla,caesar2022nuplan,li2021metadrive,kothari2021drivergym} provide reactive agents in traffic but are not designed for highly interactive corner cases where data-driven or rule-based planners cannot solve. Previous works on crowd simulation and benchmarks~\cite{crowd_Piccoli_2009} focus more on the scalability of simulating agents~\cite{Toll2015, roland2012}, and grouping in the crowd~\cite{2009crowdgroup, mahato2017particle}. We hope our framework can be complementary to these benchmarks so that one could add our agents in existing social navigation benchmarks as reactive agent planners; general navigation algorithms can be tested in our selected highly interactive environments.


\section{Conclusion}
\label{sec:conclusion}
\textbf{Conclusion and future works:}
In this project, we proposed a distributed imagined potential game framework that can generate diverse and realistic interactions in a controllable manner in various indoor environments. We show generated interactions in representative scenarios and use the distributed reactive agents to build an environment for interactive navigation training. Future works would include utilizing safe RL and exploration algorithms to learn generalized policies and conduct experiments on real robots. 



\clearpage


\bibliographystyle{IEEEtran}
\bibliography{example}	 

\clearpage

\begin{appendices}
\section{Potential Game}
\label{appendix:potential_game}
\subsection{Formulation}
We first introduce the definitions of dynamic games ($N$ agents with horizon $T$) from previous works~\cite{pilqr}. We describe a differential game by the compact notation of $\Gamma_{x_0}^T=(N, \{U_i\}_{i=1}^N, \{J_i\}_{i=1}^N, \{f_i\}_{i=1}^N)$, where $x_0$ is the initial states of all agents, and each agent seeks to optimize its cost $J_i$ under the dynamic $f_i$. The cost function $J_i(x(0), U)=S_i(x(T)) + \sum_{k=0}^{T-1}L_i(x(k), U(k))$ consists of running cost $L_i$ and terminal cost $S_i$. We look for the Nash equilibrium solution of the dynamic game defined by:
\begin{definition}
Given a differential game $\Gamma_{x_0}^T=(N, \{U_i\}_{i=1}^N, \{J_i\}_{i=1}^N, \{f_i\}_{i=1}^N)$, control signal set $U$ is an open-loop Nash equilibrium if, for $i\in [1,...,N]$:
\begin{equation}
    J_i(x_0, u^*) \leq J_i(x_0, u_i, u_{-i}^*)
\end{equation}
\end{definition}

At a Nash equilibrium, no agent has the incentive to change its current control input $u_i^*$ as such a change would not yield any benefits, given that all other agents' controls $u_{-i}^*$ remain fixed. While this equilibrium solution can best represent the cooperative multi-agent behavior in the interaction, finding the Nash equilibrium solution is challenging since there are $N$ coupled optimal control problems to solve simultaneously. Recent progress in solving this problem, especially in robotics applications, find efficient solutions to the problems under certain conditions. As proved in \cite{pilqr}, problems in a \textit{potential differential game} form can be solved by formulating a single centralized optimal control problem. We summarize their main result in the following theorem.

\begin{theorem}
\label{thm:pg}
For a given differential game $\Gamma_{x_0}^T=(N, \{U_i\}_{i=1}^N, \{J_i\}_{i=1}^N, \{f_i\}_{i=1}^N)$, if for each agent $i$, the running cost and terminal cost functions have the structure of
\begin{equation}
    L_i(x(k), u(k))=p(x(k), u(k))+c_i(x_{-i}(k), u_{-i}(k))
\end{equation}
\begin{equation}
    S_i\left(x(T)\right) = \Bar{s}(x(T)) + s_i(x_{-i}(T))
\end{equation}
then the open-loop Nash equilibria can be found by solving the following optimal control problem
\begin{equation}
\begin{aligned}
\min_{U} \quad & \sum_{k=1}^{T-1}p(x(k), u(k))+\Bar{s}(x(T))\\
\textrm{s.t.} \quad & 
x_i(k+1) = f_i(x_i(k), u_i(k)) \\
\end{aligned}
\label{eq:pgeq}
\end{equation}
\end{theorem}

The key takeaway from this theorem is that one can formulate a differentiable potential game if all the cost function terms can be decomposed into potential functions ($p(\cdot), \Bar{s}(\cdot)$) that depend on the full state and control vectors of all the agents, and other cost terms ($c_i(\cdot), s_i(\cdot)$) that have no dependence on the state and control input of agent $i$. Then the optimal solution for both agents can be solved by the centralized problem Eq.\ref{eq:pgeq} using only $p, \Bar{s}$.

\subsection{System Notations and Assumptions}
To simplify the problem, we make several assumptions on system dynamics. 
We assume that all agents are modeled using the same unicycle dynamic model. The state vector $x_i = [p_{x, i}, p_{y, i}, \theta_i, v_i]$, the control vector $u_i = [a_i, w_i]$. 
The discrete-time dynamic equations of the system are: 

\begin{equation}
\begin{aligned}
    p_{x,i}(k+1) &= p_{x,i}(k) + T_s \ v_i(k) \ cos(\theta_i(k)) \\ 
    p_{y,i}(k+1) &= p_{y,i}(k) + T_s \ v_i(k) \ sin(\theta_i(k)) \\ 
    \theta_i(k+1) &= \theta_i(k) + T_s \ w_i(k) \\ 
    v_i(k+1) &= v_i(k) + T_s \ a_i(k)
\end{aligned}
\end{equation}

System notations are summarized in Table \ref{tab: notation}.

\begin{table}[h]
\centering
\caption{Notation describing common variable.}
\label{tab: notation}
\begin{tabular}{l l || l l }
    \Xhline{1pt}
     & Definition & & Definition (Default value)\\
    \hline
     $p_x, p_y$ & position in 2D & Q & state weight ([0.01, 0.01, 0, 0])\\
     $\theta$ & heading angle & R & input weight ([1, 1])\\
     v & velocity & D & safety weight (40) \\
     a & acceleration & B & back up weight (10)\\
     w & angular velocity & r & safety radius (1.2$\sim$2.0)\\
     $O$ & static obstacles & $T_s$ & sampling time (0.1) \\
    \Xhline{1pt}
\vspace{-0.15in}

\end{tabular}
\vspace{-0.1in}
\end{table}

\section{IPG Implementation}
\label{appendix:implementation}
\subsection{Cost functions with Potential Game Formulation}

Based on Theorem \ref{thm:pg}, Nash equilibrium solutions of all agents can be solved by Eq. \ref{eq:pgeq} if the interaction can be formulated into a potential game. Here, we show the running cost function $L_i(x)$ in the potential game, consisting of the stage cost term  $C_{tr,i}^{0:T-1}(x_i, u_i)$,  the collision avoidance term $C_{a,ij}(x_i, x_j)$ and the reverse avoidance term $C_{b,i}(x_i)$. The stage cost includes the minimum goal distance and inputs penalty terms using the current state and input:
\begin{equation}
\begin{aligned} 
C_{tr,i}^{0:T-1}(x_i, u_i) = (x_i - g_i)^{\intercal}Q_i(x_i - g_i) + u_i^{\intercal} R_i u_i
\end{aligned}
\end{equation}
The collision avoidance term is counted when the distance between two agents $d_{ij}$ is smaller than the safety distance:
\begin{equation}
    C_{a, ij}(x_i, x_j) =  \begin{cases} (d_{ij} - d_{safe})^2 \cdot D_i, & \text{if $d_{ij} < d_{safe}$} \\ 0, & \text{others} \end{cases}
\end{equation}
and satisfy the symmetric property $C_{a, ij}(x_i, x_j) = C_{a, ji}(x_j, x_i)$ for potential game described in \cite{pilqr}.

The reverse avoidance term discourages the agent for moving backward:

\begin{equation}
C_{b, i}(x_i) =  \begin{cases} |v_i| \cdot B_i, & \text{if $v_i < 0$} \\ 0, & \text{others} \end{cases}
\end{equation}



To prove this running cost function is a potential game, we can represent the $p(x, u)$ and $c_i(x_{-i}, u_{-i})$ in Theorem \ref{thm:pg}:
\begin{equation}
\begin{aligned} 
p(x, u) = &\sum_{i=1}^N C_{tr, i}^{0:T-1}(x_i, u_i) \\ 
& + \sum_{1 \leq i < j} C_{a, ij}(x_i, x_j) + \sum_{i=1}^N C_{b, i}(x_i)\\
\end{aligned}
\end{equation}
\begin{equation}
\begin{aligned} 
c_i(x_{-i}, u_{-i}) = 
&-\sum_{j \neq i} C_{tr, j}^{0:T-1}(x_j, u_j) \\
&- \sum_{\substack{1 \leq j < k \\ j, k \neq i}} C_{a, jk}(x_j, x_k) - \sum_{j \neq i} C_{b, j}(x_j)
\end{aligned}
\end{equation}
With this representation, we show that the running cost $L_i(x_i, u_i)= p(x, u) + c_i(x_{-i}, u_{-i})$ follows the Theorem \ref{thm:pg}, 

Similarly, the terminal cost $S_i(x(T)) = C_{tr,i}^T(x_i, u_i) = \bar{s}(x(T)) + s_i(x_{-i}(T))$, with terminal cost term:
\begin{equation}
\begin{aligned} 
C_{tr,i}^T(x_i, u_i) = (x_i(T) - g_i)^{\intercal}Q_i(x_i(T) - g_i)
\end{aligned}
\end{equation}
Set $\bar{s}(x(T))$ and $s_i(x_{-i}(T))$ to be :
\begin{equation}
\begin{aligned} 
&\bar{s}(x(T) = \sum_{i=1}^N C_{tr, i}^T(x_i),s_i(x_{-i}(T)) = -\sum_{j \neq i}^N C_{tr, j}^T(x_j)
\end{aligned}
\end{equation}
    We have the required terminal cost $S_i(x(T))$ in Theorem \ref{thm:pg}. 

To address the problem in scenarios involving obstacles, we introduce some extra constraints in addition to the existing dynamic constraints, including the state boundary constraint, input constraint, and obstacle avoidance constraint. These constraints can be added as weighted costs into $J_i$ and don't affect the potential game assumptions. 

One important difference between centralized planning and the IPG setting is the safety distance $d_{safe}$. For centralized planning, the maximum safety distance between them is used $d_{safe} = max(r_i, r_j)$, for IPG, each agent assumes others have the same safety distance, $d_{safe, i}=r_i$, unless it changes its estimation. 

The full IPG problem for each agent to solve is:
\begin{equation}
\label{eq: ipg}
\begin{aligned} 
    \min_{U}  \ &\sum_{k=1}^{T-1} p(x(k), u(k)) + \bar{s}(x(T))\\ 
    s.t. \quad &x(k+1) = f(x(k), u(k)) \\ 
    &x(0) = x_0 \\ 
    &x_L \leq x(k) \leq x_U \\ 
    &u_L \leq u(k) \leq u_U \\ 
    &r_{obs} - dis(x(k), O_m) \leq 0, m=1...M \\ 
    &r_{i} - dis(x_i(k), x_j(k)) \leq 0, i, j=1...N\\
\end{aligned}
\end{equation}
where $x_U, x_L, u_U, u_L$ are the state and input boundaries, and $r_{obs}$ is the radius of the circle obstacle.

\begin{figure*}[t]
    \centering
\includegraphics[width=1.0\textwidth]{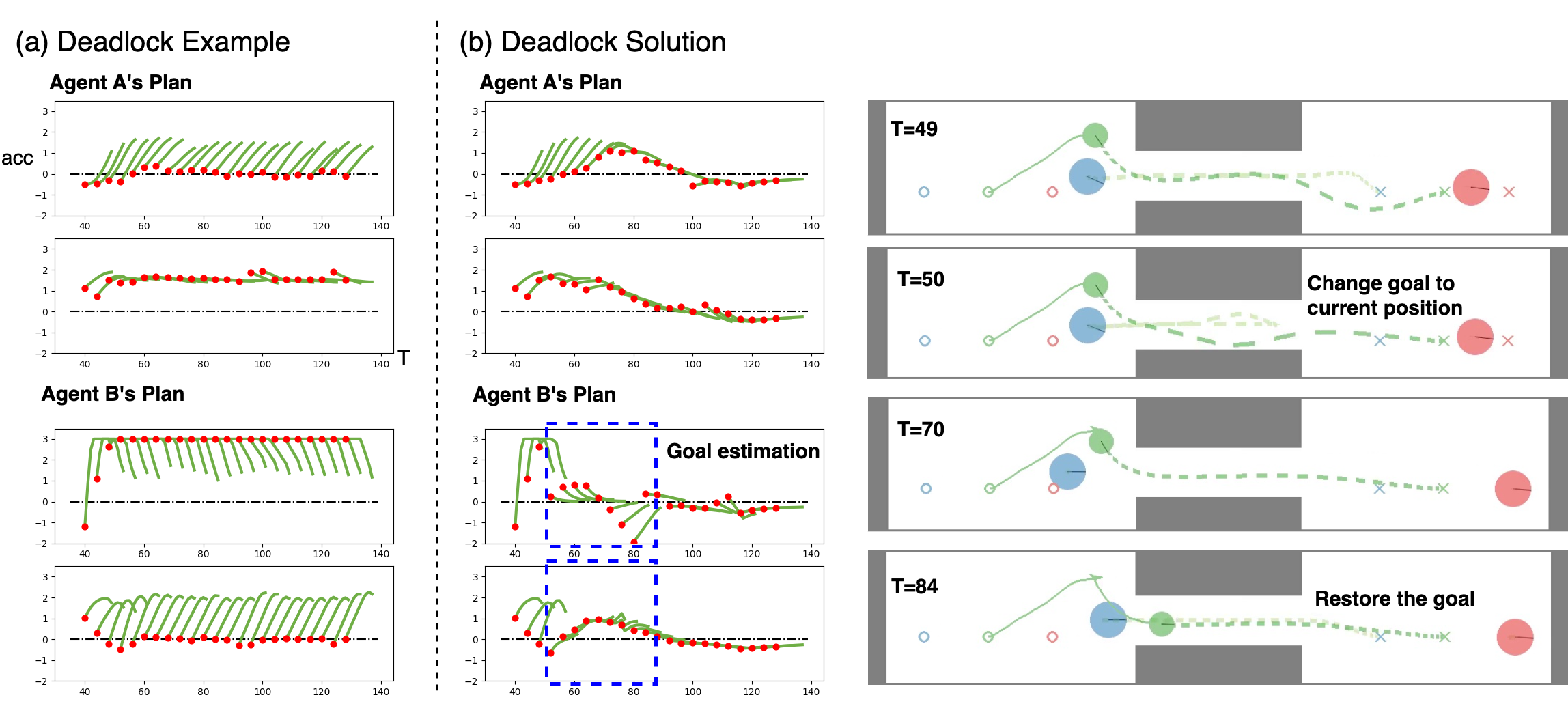}
    \caption{Deadlock example in HallWay scenario and its solution. We plot the acceleration vs. time diagram where green lines are the planned first 10 actions, and the red points are the executed actions in the next step. Without goal estimation, agents in conflict plan on yielding, resulting in a deadlock. However, with goal estimation, Agent B perceives that Agent A is yielding, though Agent A is expected to move in Agent B's game. Consequently, Agent B modifies Agent A's goal to Agent A's current position. If Agent A begins to move, contradicting the initial assumption of yielding, Agent B will revert Agent A's goal to its original state in Agent B's model. This implementation of goal estimation allows the plans of Agent A and Agent B to align more closely with the executed actions.}
    \label{fig:deadlock solving}
\end{figure*}

\subsection{Solutions to deadlocks}
In certain scenarios involving deadlocks, consider a two-agent deadlock as an example: both agents plan to remain stationary, each expecting the other to move first. Consequently, a deadlock arises because each agent yields to the other. One strategy to address this issue involves increasing the ego-agent's safety distance to enforce yielding behaviors within their imaged potential game, as proposed by \cite{sun2024distributed}, an example is shown in Figure 3(b) and discussed in Section 3.2 in the paper.

Alternatively, another method of resolving the deadlock is to modify the goal position to induce yielding behavior. 
Specifically, the goal position of the other agent is set to its current location. This setup effectively changes the immediate objective of the other agent to yielding since that is the most effective way to solve the interaction. We have observed that this approach successfully resolves such deadlocks, as demonstrated in Figure \ref{fig:deadlock solving}. However, in practice, since all agents are making independent decisions, there is generally no guarantee for solving deadlocks since agents can follow the same deadlock-resolving strategies, which causes another round of deadlock.

\subsection{Practical Implementation Details}
In this section, we introduce several practical techniques used in our IPG simulator. Specifically, we present three key techniques: Multiple Solution, Adaptive Safety Penalty, and Conservative Assumption, which are designed to improve safety and avoid deadlocks. We tested these techniques in 100 randomly generated three-agent Hallway scenarios and evaluated them based on Success, Collision, and Truncated (where agents fail to reach their goals within the total time step). We set the total time step to 400, which is an extremely long duration for agents to reach their goals. The results in Table \ref{tab: practical techs} show that these three techniques improve the success rate of our simulator.

\begin{table}[ht]
\centering
\begin{tabular}{l c c c}
\hline
\textbf{}               & \textbf{Success(\%)} & \textbf{Collision(\%)} & \textbf{Truncated(\%)} \\ 
\hline
Ours                   & 100              & 0                  & 0                  \\ 
-Multiple Solution                  & 29               & 71                 & 0                  \\ 
-Adaptive Penalty            & 97               & 2                  & 1                  \\ 
-Conservative       & 94               & 6                  & 0                  \\ 
\hline
\end{tabular}
\caption{Success Rate in 100 random three-agent HallWay scenarios.}
\label{tab: practical techs}
\end{table}

\subsubsection{Multiple Solutions}
\begin{figure*}[t]
    \centering
\includegraphics[width=1.0\textwidth]{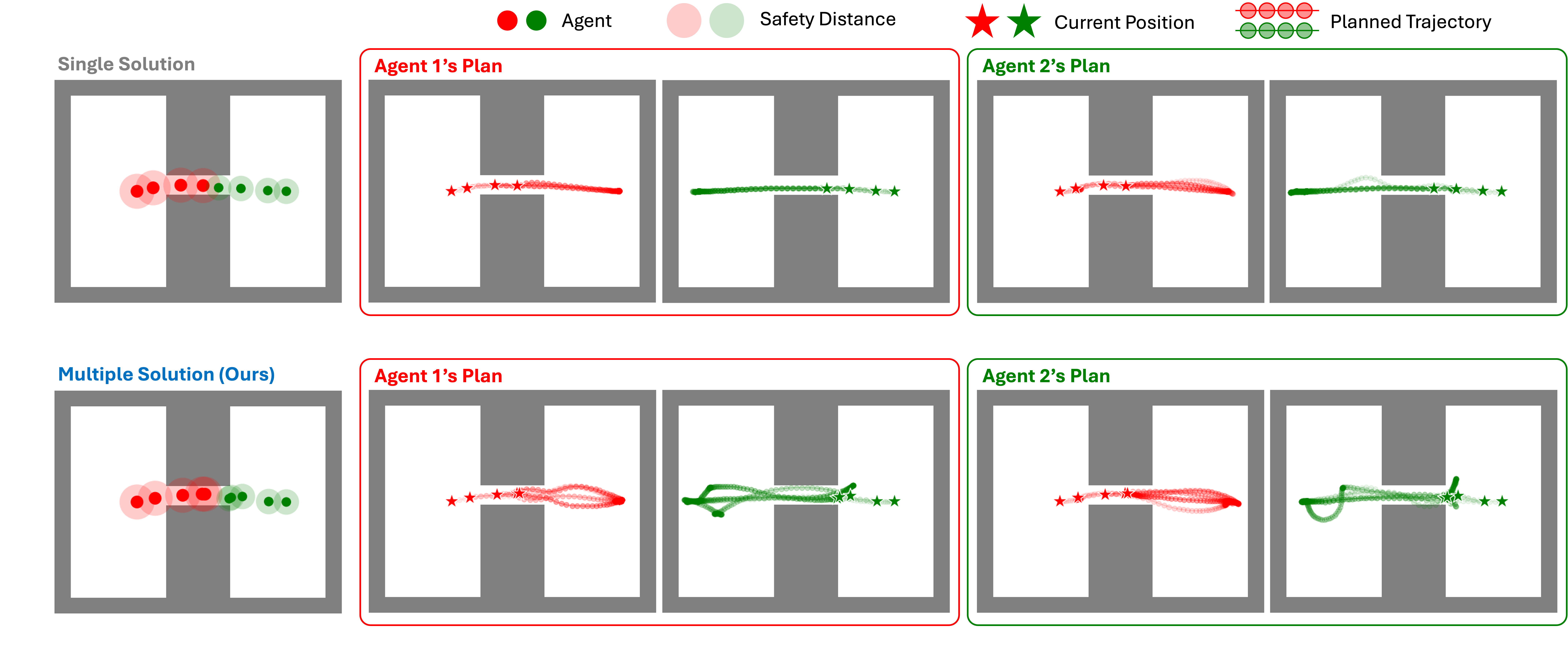}
    \caption{Example of Single Solution vs. Multiple Solutions: In the Single Solution approach, Agent 1 and Agent 2 will collide. We can observe that their plans remain consistent across different time steps, showing that the iLQR to repeatedly converge to the same infeasible point within a limited number of iterations. In contrast, with the Multiple Solutions approach, the two agents slow down when they are close to each other. Their plans remain consistent in the initial time steps but diverge significantly as they approach one another, providing feasible solutions for both agents.}
    \label{fig: multi sol}
\end{figure*}

To solve Problem \ref{eq: ipg}, we employ the iLQR method using trajax. If the previous solution is feasible, we use it as the initialization. If it is not feasible, we use a zero initialization. Since this approach only generates a single solution, we refer to it as the "Single Solution" strategy. However, this strategy has several drawbacks. The main issue is that iLQR may converge to the same point within a limited number of optimization steps, and this point could be infeasible. As shown in Figure \ref{fig: multi sol}, when using the Single Solution strategy, both agent 1 and agent 2 converge to similar infeasible solutions, leading to a collision.

To address this, we generate multiple solutions using different initialization methods and choose one with the lowest cost, which we refer to as the "Multiple Solution" strategy. Specifically, we retain the original initialization strategy and introduce the following additional strategies:
\begin{itemize}
    \item Gaussian noise. The mean of the Gaussian distribution is randomly selected to bias actions toward acceleration, deceleration, or turning left or right, resulting in diverse and reasonable behaviors rather than random walks.
    \item Single-agent solution. For each agent, we first plan a single-agent solution without considering other agents and use that solution as the initialization.
\end{itemize}

As shown in Figure \ref{fig: multi sol}, with the Multiple Solution strategy, the two agents can generate different solutions when they are close to each other and then slow down to avoid collisions. Additionally, we use multithreading to accelerate the calculation of multiple solutions.

\subsubsection{Adaptive Safety Penalty}
\begin{figure*}[t]
    \centering
\includegraphics[width=1.0\textwidth]{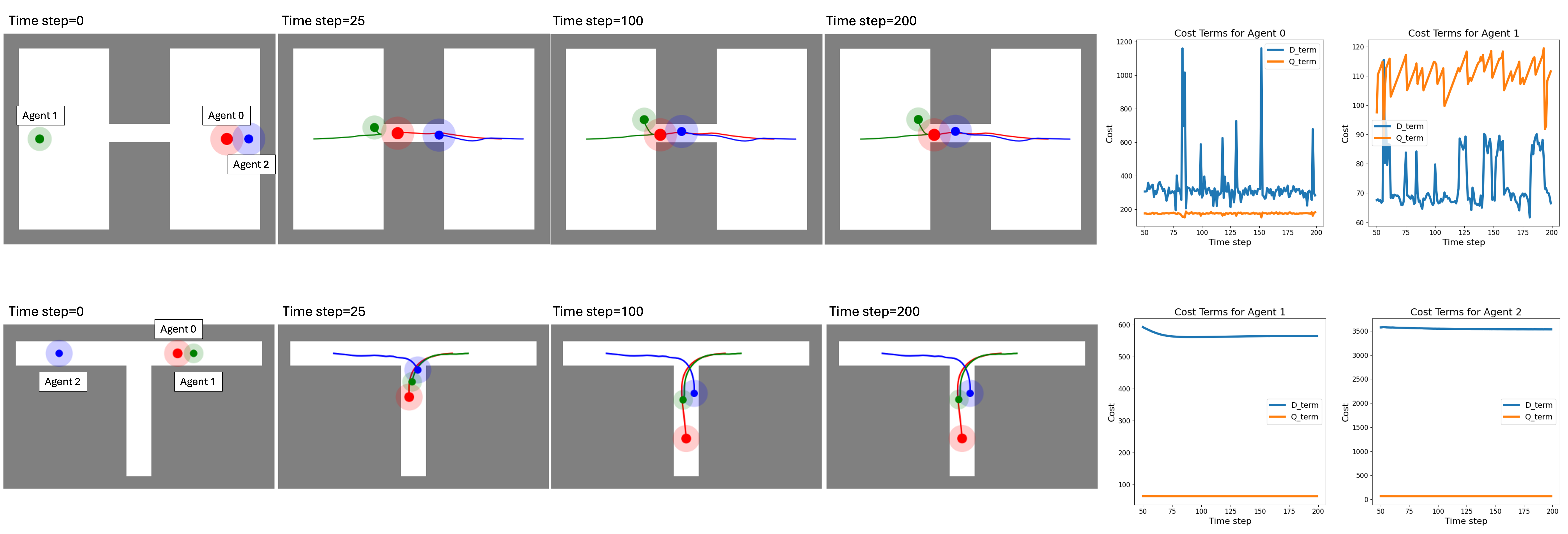}
    \caption{Deadlock due to the safety penalty in the Hallway (Top) and T-Intersection (Bottom) scenarios. We plot the Safety Penalty (D\_term) and Goal Reach Cost (Q\_term) during the deadlock. It is evident that the safety penalty is significantly larger than the goal-reaching cost, indicating that the agents prefer to remain stationary to maintain a safe distance from other agents, rather than sacrificing safety (no collision) to reach their goals. In the Hallway scenario, Agent 0 could break the deadlock by slightly moving left and down, but the high D\_term (which exceeds the Q\_term) prevents this. Agent 1 has comparable Q\_term and D\_term values. In the T-Intersection scenario, both Agent 1 and Agent 2 have much higher D\_term values than Q\_term, which prevents them from reaching their respective goals.}
    \label{fig: safety penalty}
\end{figure*}
The constant safety penalty term $C_{a, ij}(x_i, x_j)$ (denoted as D\_term) can sometimes prevent agents from reaching their goals. This occurs when $\sum_{j\neq i}C_{a, ij}(x_i, x_j) \gg (x_i - g_i)^{\intercal}Q_i(x_i - g_i)$, as illustrated in Figure \ref{fig: safety penalty}. To mitigate this issue, we propose a strategy: when traffic flow speed is low, the safety penalty should also be low (for example, when parking a car); when traffic flow speed is high, the safety penalty should be much larger (for example, when driving on a highway). Based on this, we adjust $D_i$ or $d_{safe}$ according to $\text{min}{v_i}_{i=1}^N$. However, ensuring that $\sum_{j\neq i}C_{a, ij}(x_i, x_j) \leq (x_i - g_i)^{\intercal}Q_i(x_i - g_i)$ does not guarantee that the gradient will guide the agent toward the goal. Therefore, in our implementation, we chose to adjust $d_{safe}$. Although we plan for $T$ steps and $d_{safe}$ may vary across different time steps, we execute only the next step at a time. Thus, we use a constant $d_{safe}$. We also experimented with designing a function $d_{safe} = f(\text{min}\{v_i\}_{i=1}^N)$, but found that this offered no benefits for the solutions but introduced significant computational complexity and latency due to the increased complexity of the cost function.

\subsubsection{Conservative Assumption}
Collisions pose a significant challenge in highly interactive scenarios, especially in distributed settings. Collisions occur when the solutions among different agents are not well-aligned, so when agents are close to each other, they lack sufficient safety tolerance to avoid collisions. To address this, we designed a strategy where the movement changes of other agents are limited, inspired by the way humans expect others not to make drastic changes in their movements when planning their own. Specifically, we reduce the control input limits of other agents, i.e., the absolute values of $u_L$ and $u_U$.

\section{Reinforcement Learning Experiments}
\label{app: RL results}

\begin{figure*}[t]
    \centering
\includegraphics[width=0.9\textwidth]{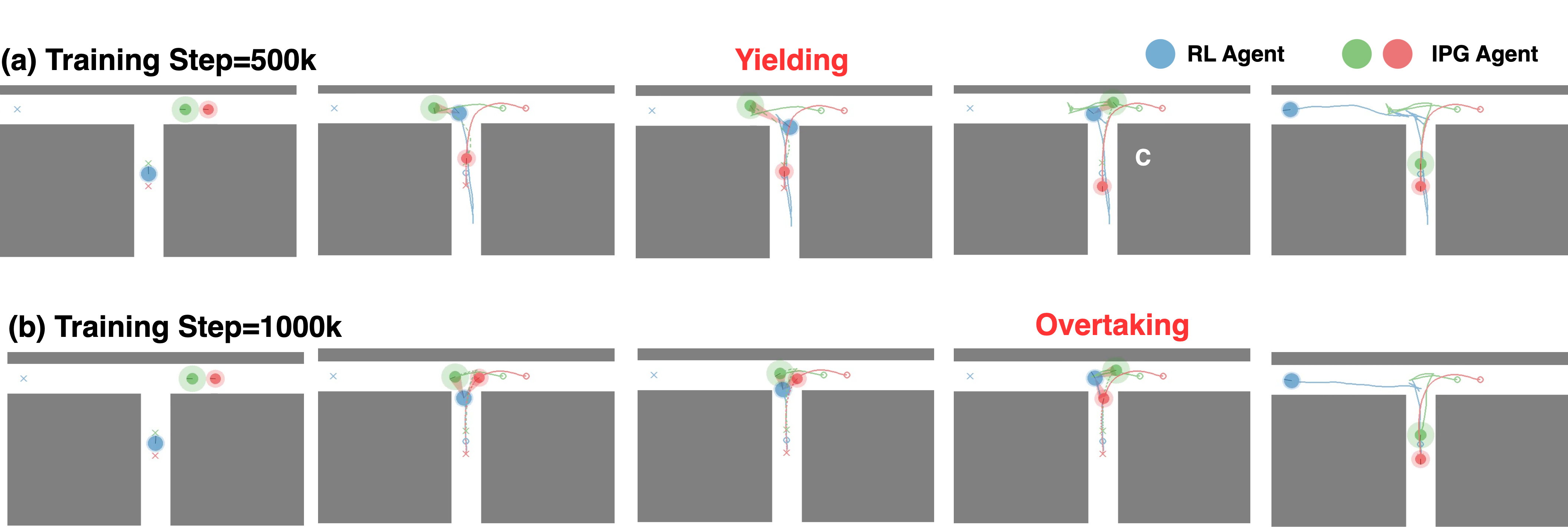}
    \caption{The behavior of reinforcement learning (RL) agents at various stages of training exhibits notable changes. As training progresses, there is a discernible increase in the aggressiveness of the RL agent's behavior, predicated on the assumption that the IPG agent will yield.}
    \label{fig:rl results}
\end{figure*}

\subsection{Gym Environment settings}
\label{appendix:gym}
\paragraph{Observations} 
We set that the observation of the reinforcement learning (RL) agent contains the environment, its own current state, and its goal point. When it detects other agents (as detailed in Section 3.1 on Occlusions), the agent also acquires the current state and goal point of the observed agents. 
For fixed-goal in a fixed-environment setting, since the environment and the goal points of both the RL and observed agents are static, there is a risk that the neural network may ignore such crucial information. Therefore, we adopt an agent-centric setting to give a more generalized representation of observation: expressing the observed data relative to the frame of reference of the RL agent. Specifically, 
\begin{itemize}
    \item \textbf{The state of RL agent}. Denote the state of one agent in the environment as $s=(x,y,\theta,v)$ where $\theta$ is the heading angle and $v$ is the absolute velocity. Denote the goal as $s_g=(x_d,y_d,\theta_d,v_d)$. Thus, the distance to goal $d$ is $\sqrt{(x-x_d)^2+(y-y_d)^2}$. The state of the RL agent is represented as $(x_g-x,y_g-y,\theta_g-\theta,\cos (\theta_g-\theta), \sin (\theta_g-\theta),v_g,d)$.
    \item \textbf{The states of the observed agents}. Denote the state of one observed agent $i$ as $(x^i,y^i,\theta^i,v^i)$ and its goal as $(x_d^i,y_d^i,\theta_d^i,v_d^i)$. The distance to goal $d^i$ is $\sqrt{(x^i-x^i_d)^2+(y^i-y^i_d)^2}$. Its state is represented as $(x^i-x,x^i-y,\theta^i-\theta,\cos (\theta^i-\theta), \sin (\theta^i-\theta),v^i,x_g^i-x,y^i_g-y,\theta^i_g-\theta,\cos (\theta^i_g-\theta), \sin (\theta^i_g-\theta),v^i_g,d^i)$.
    \item \textbf{Environment boundary}. Denote the boundary as $(x_{min}, x_{max}, y_{min},y_{max})$. The state of the boundary is represented as $(x-x_{min}, x-x_{max}, y-y_{min}, y-y_{max} )$.
    \item \textbf{Rectangular obstacle}. Denote the information of the rectangular obstacle is $(o_x,o_y,h,w)$ where $(o_x,o_y)$ are the location of its center, $h$ and $w$ are height and width. It is represented as $(x-(o_x+w/2), x-(o_x-w/2), y-(o_y+h/2), y-(o_y-h/2))$. 
\end{itemize}

\paragraph{Actions} The acceleration and the angle velocity of the RL agent. 
\subsection{Reward function}
\label{appendix:reward}
Designing rewards for interactive agents in these scenarios is hard; here we share some of the insights we found during experiments:
\begin{itemize}
    \item Sparse reward is hard to learn since the interaction is long-horizon. We use the goal-reaching reward to provide dense reward supervision when the agent is closer to the goal. 
    \item We find if $0<r_g<|r_e|$, the agent will choose not to move towards the goal and get stuck. It can be solved when we set a desired velocity, but in highly interactive navigation tasks, yielding (staying still) is always the case. Setting the desired velocity will lead to abnormal behaviors. Thus, we use the hierarchy reward setting, which means if the agent moves toward the goal, goal reaching reward should be greater than the energy cost. If the $r_g>0$,  $\min (r_g+r_e, 0.01)$.
    \item We cannot set the collision penalty too high in our environments. If the penalty is too high, it might encourage the agent to avoid an enlarged region of the obstacles. In the HallWay scenario, an enlarged region of obstacles can easily cover the hallway, so the agent will directly avoid entering the hallway so that it is hard to approach the goal (the agent needs first to enter the hallway, then reach the goal on the opposite side of the hallway). However, too small collision avoidance will lead to the case that the agent chooses to make collisions to avoid the accumulated negative rewards ($r_g+r_e$). It is serious in the social navigation task because it is normal that the agent needs to move backward to yield to the others (let the others pass the single-file hallway first) and get very large negative rewards ($r_g+r_e$). To avoid this happening, we set the survival reward and the highest reward level. If the agent is alive, its minimal reward is 0.01. 
\end{itemize}

\subsection{Example of a converged RL agent}
We train a reinforcement learning (RL) agent within a T-intersection environment, incorporating two Imaged Potential Game (IPG) agents, each with a fixed goal and initial position. We employ the Proximal Policy Optimization (PPO) algorithm \cite{schulman2017proximal} in Stable Baseline3 \cite{raffin2021stable} with default settings for training purposes. The training process is conducted across 16 parallel gym environments and for 1,000,000 environment steps.

The illustration in Figure \ref{fig:rl results} demonstrates that the reinforcement learning (RL) agent can emulate human-like behaviors, including yielding and overtaking. Significantly, within the framework of the current reward design, there is a noticeable enhancement in the aggressiveness of the RL agent's behavior as training continues. This observation is pivotal for evaluating the effectiveness of a policy and its corresponding reward design in highly interactive scenarios.

For more examples of interactions generated interactions under different settings:
\begin{itemize}
    \item Multiple IPG agents interacting in different scenarios
    \item IPG agent interacting with non-collaborative agents
    \item IPG agent interacting with RL agents (success and failures)
\end{itemize}
we encourage visiting our website~\cite{website}. We leave finding effective reinforcement learning methods to learn reactive policies that can collaborate and navigate in the same policy to our future work.
\end{appendices}

\end{document}